\documentclass[11pt,a4paper]{article}
\usepackage{times,latexsym}
\usepackage{url}
\usepackage[T1]{fontenc}

   \usepackage[acceptedWithA]{tacl2021v1}
\usepackage{xcolor}
\newcommand{\ensuretext}[1]{#1}

\newcommand{\samarker}{\ensuretext{\textcolor{orange}{\ensuremath{^{\textsc{S}}_{\textsc{A}}}}}}
\newcommand{\rxmarker}{\ensuretext{\textcolor{purple}{\ensuremath{^{\textsc{R}}_{\textsc{X}}}}}}

\newcommand{\mycomment}[3]{}

\newcommand{\sa}[1]{\mycomment{\samarker}{#1}{orange}}
\newcommand{\rx}[1]{\mycomment{\rxmarker}{#1}{purple}}
\newcommand{\ignore}[1]{}

\usepackage[inline]{enumitem}
\usepackage[utf8]{inputenc} 
\usepackage[T1]{fontenc}    
\usepackage[english]{babel}
\usepackage{hyperref}       
\usepackage{url}            
\usepackage{amsfonts}       
\usepackage{nicefrac}       
\usepackage{microtype}      
\usepackage{booktabs}
\usepackage{amsfonts}
\usepackage{amsmath}
\usepackage{multirow}
\usepackage{graphicx}
\usepackage{enumitem}
\usepackage{subcaption}
\usepackage{caption}
\usepackage{rotating}
\usepackage[normalem]{ulem}
\usepackage{amssymb}
\usepackage{colortbl}
\usepackage{linguex}

\usepackage{siunitx} 
\usepackage{multirow, makecell}
\usepackage{pifont}
\newcommand{\cmark}{\ding{51}}%
\newcommand{\xmark}{\ding{55}}%
\usepackage{siunitx} 
\usepackage{tabularx}
\usepackage[all]{nowidow}
\usepackage{colortbl}
\usepackage{nicematrix,tikz}
\usepackage{xcolor}

\definecolor{hotpink}{HTML}{EF7C8E}
\definecolor{tiffanyblue}{HTML}{A0E7E5}
\definecolor{mint}{HTML}{B4F8C8}
\definecolor{paleyellow}{HTML}{FBE7C6}
\definecolor{rosewater}{HTML}{D8A7B1}
\definecolor{cream}{HTML}{FAE8E0}

\newcommand{\ts}{\textsc{TS}\xspace}
\newcommand{\nlp}{\textsc{NLP}\xspace}

\newcommand{\bleu}{\textsc{Bleu}\xspace}

\newcommand{\sari}{\textsc{SARI}\xspace}

\newcommand{\kis}{\textsc{KIS}\xspace}
\newcommand{\wiki}{\textsc{ControlT5-Wiki}\xspace}
\newcommand{\musssup}{\textsc{MUSS-Sup}\xspace}
\newcommand{\mussunsup}{\textsc{MUSS-Unsup}\xspace}
\newcommand{\editnar}{\textsc{EditCL}\xspace}
\newcommand{\controlsup}{\textsc{ControlSup}\xspace}
\newcommand{\chatgpt}{\textsc{ChatGPT}\xspace}
\newcommand{\correctness}{adequacy\xspace}

\usepackage{enumitem}
\setlist{nolistsep,leftmargin=*}

\title{Do Text Simplification Systems Preserve Meaning? \\ A Human Evaluation via Reading Comprehension}

\author{Sweta Agrawal\thanks{\quad Work done while at the University of Maryland.}\\
  Instituto de Telecomunicações \\
      {\tt swetaagrawal20@gmail.com} \\\And
  Marine Carpuat \\
  University of Maryland \\
  {\tt marine@umd.edu} \\}

\begin{document}
\maketitle
\begin{abstract}
Automatic text simplification (\ts) aims to automate the process of rewriting text to make it easier for people to read. A pre-requisite for \ts to be useful is that it should convey information that is consistent with the meaning of the original text. However, current \ts evaluation protocols assess system outputs for simplicity and meaning preservation without regard for the document context in which output sentences occur and for how people understand them. In this work, we introduce a human evaluation framework to assess whether simplified texts preserve meaning using reading comprehension questions.  With this framework, we conduct a thorough human evaluation of texts by humans and by nine automatic systems. Supervised systems that leverage pre-training knowledge achieve the highest scores on the reading comprehension (RC) tasks amongst the automatic controllable \ts systems. However, even the best-performing supervised system struggles with at least 14\% of the questions, marking them as ``unanswerable'' based on simplified content. We further investigate how existing \ts evaluation metrics and automatic question-answering systems approximate the human judgments we obtained. 

\end{abstract}

\section{Introduction}

Rewriting text so that it is easier to understand has the potential to help a wide range of audiences including non-native speakers \cite{SarahPetersenMariOstendorf2007, DavidAllen2009, crossley2014s}, children \cite{watanabe2009facilita} or people with reading or cognitive disabilities \cite{AlonzoSeitaGlasserHuenerfauth2020Automatic} access information more easily \citep{chandrasekar-etal-1996-motivations, stajner-2021-automatic}. Online resources such as the Newsela \cite{news} and OneStopEnglish \cite{onestop} or the Cochrane systematic reviews \cite{cochrane}, provide text articles simplified by human editors so that they are easier to understand by K-12 students, English speakers with limited proficiency, and lay people seeking to understand medical literature respectively. This has motivated a wealth of Natural Language Processing research on text simplification, framed as the task of rewriting an input text into a simpler version while preserving the core meaning of the original \citep{chandrasekar1997automatic}, which has been addressed with approaches ranging from dedicated supervised systems \citep{specia2010translating, zhang2017sentence, scarton-specia-2018-learning, martin2020controllable, jiang2020neural, devaraj-etal-2021-paragraph, sheang-saggion-2021-controllable, agrawal-carpuat-2022-imitation, martin-etal-2022-muss} to prompting black-box pre-trained models \citep{feng2023sentence, kew-etal-2023-bless}.

However, texts that are easier to read are not helpful if they mislead readers by providing information that is not consistent with the original document. This can happen with automatic \ts outputs where deletions or inaccurate rewrites can change how a text is understood \cite{devaraj-etal-2022-evaluating}.
Assessing to what extent the meaning of the original text is preserved should therefore be a critical dimension of \ts evaluation \cite{stajner-2021-automatic}, and a pre-requisite to determining whether and how \ts can be used in practice. Additionally, evaluating individual sentences out of context may not be sufficient to establish whether model-generated texts preserve meaning, as human simplifications often occur at the document or the passage level \cite{devaraj-etal-2022-evaluating}. Yet, \ts outputs are primarily evaluated intrinsically, with automatic metrics that compare system outputs with human-written reference simplifications and/or the original source \citep{papineni2002bleu, XuNapolesPavlickChenCallison-Burch2016, maddela2022lens}, or with generic human assessments of simplicity and meaning preservation of individual sentences outside of a context of use \citep{schwarzer2018human}. While these evaluations can guide model development, they do not address whether readers get information from the simplified text that is consistent with the original content.

\begin{figure*}[htb!]
\centering
\begin{subfigure}{0.97\linewidth}
  \centering
\includegraphics[width=\textwidth]{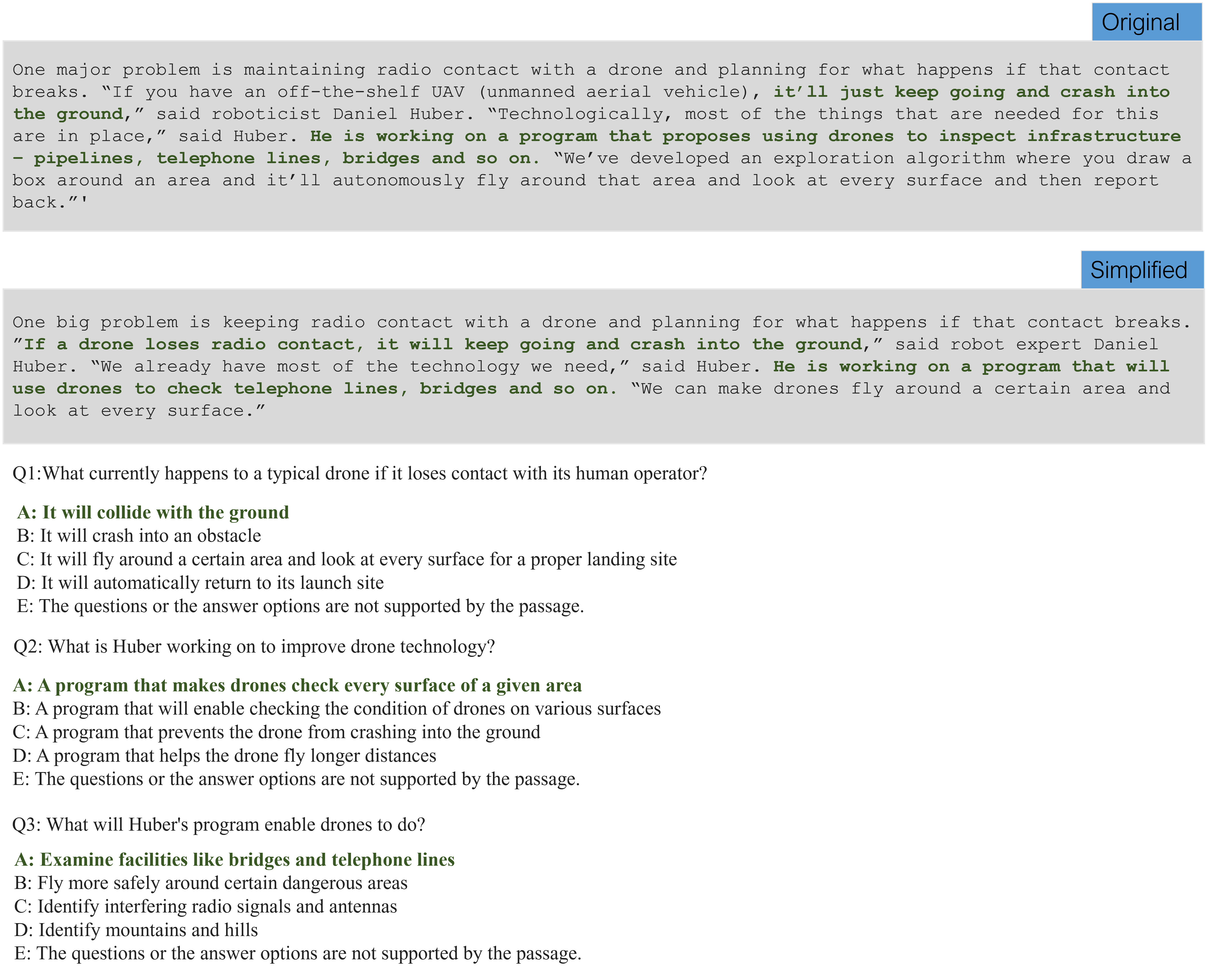}
\end{subfigure}%
 \caption{ RC questions to be answered after reading either the original or the simplified text. The answer options include the correct answer (A), three incorrect options of varying difficulty (B,C,D), and an option (E) that captures questions rendered unanswerable after automatic \ts.}\label{fig:example_qa}
 \vspace{-0.5cm}
\end{figure*}

In this work, we conduct a human evaluation of the ability of state-of-the-art \ts systems to preserve the meaning of the original text by measuring how well people can answer questions about key facts from the original text after reading a simplified version. We design RC tasks to directly assess meaning preservation in \ts, different from prior uses of reading comprehension to assess people's reading efficiency  \citep{angrosh-etal-2014-lexico, laban-etal-2021-keep}. This framework lets us conduct a controlled comparison of simplified texts, whether written by humans or by \ts systems: we compare people's ability to answer questions about the original text, a simplified version written by humans, and nine \ts-generated versions that represent a diverse set of supervised and unsupervised approaches from the recent \ts literature. \footnote{Collected annotations and code are released at \url{https://github.com/sweta20/ATS-EVAL.git}}

 We first discuss relevant literature for TS evaluation and the use of RC exercises to assess simplified or other model-generated texts in Section~\ref{sec:background}. Next, Section~\ref{sec:method} elaborates on our RC-centered human evaluation framework, and  Section~\ref{sec:setting} delves into the various design choices we made. Section~\ref{sec:results} demonstrates the robustness of our evaluation and presents the main results. As we will see, supervised systems that utilize pre-training knowledge achieve the highest level of accuracy in RC tasks compared to other automatic controllable \ts systems (\S\ref{sec:correctness}). However, at least 14\% of the questions remain unanswerable even for the best-performing system due to the errors introduced by these systems (\S\ref{sec:answerability}). 
 In Section~\ref{sec:automaticmetrics}, we shift our focus towards a meta-evaluation of existing automatic TS evaluation metrics which indicates that the 3-way comparison used in SARI makes it a reliable metric for system-level evaluation at the paragraph level. Finally, we include a preliminary discussion and analysis of the potential for automating the RC-based evaluation through the application of model-based question-answering techniques in Section~\ref{sec:modelqa}.

\section{Background} \label{sec:background}

How to design human and automatic evaluation protocols for \ts is a research question onto itself. While automatic metrics are key to system development, commonly used metrics like BLEU \cite{papineni2002bleu}, SARI \cite{XuNapolesPavlickChenCallison-Burch2016} or the Flesch-Kincaid Grade Level \cite{flesch2007flesch} have low correlation with human judgments of simplicity \cite{sulem-etal-2018-bleu, alva-manchego-etal-2021-un, maddela2022lens, tanprasert-kauchak-2021-flesch}. 
This suggests that these metrics can fail to capture meaningful differences between simplified texts. Furthermore, there is no standardized framework for measuring the \textbf{adequacy} of simplified outputs \cite{stajner-2021-automatic, grabar-saggion-2022-evaluation}, where adequacy refers to the degree to which the generated text accurately conveys the meaning from the original text \cite{blatz-etal-2004-confidence}. \footnote{We use the terms ``adequacy'' and ``meaning preservation'' interchangeably to convey whether the information from the original is preserved in the simplified throughout this paper. }

Prior work highlights the importance of manually evaluating \ts systems. For instance, \citet{maddela2022lens} introduce \textsc{RANK \& RATE}, a human evaluation framework that rates simplifications from several models at the sentence level by leveraging automatically annotated edit operations that are verified by annotators.  These edit operations are then used in rating the output texts on a scale of 0 to 100. However, this rating is meant to jointly account for meaning preservation, simplicity, and fluency. \citet{devaraj-etal-2022-evaluating} show that factual errors often appear in both human and automatically generated simplified texts (at the sentence level), and define an error taxonomy to account for both the nature and severity of errors. Yet, these intrinsic evaluations do not directly tell us whether people correctly understand key facts conveyed in the original after reading a simplified text. Moreover, the evaluation is only performed at the sentence level without accounting for the context in which they appear which can impact the overall assessments as noted by \citet{devaraj-etal-2022-evaluating}.

Reading comprehension tests are standard tools used by educators to assess readers' understanding of text materials, and thus provide an assessment of \ts that is more in line with its intended use. They have been used to show that human-simplified texts are easier to comprehend by L2 learners \cite{long1993modifications,tweissi1998effects, oh2001two, crossley2014s, rets2021simplify}, as well as secondary and post-secondary students  \cite{heydari2013effectiveness}. For instance, \citet{long1993modifications} conduct a reading comprehension study with 483 Japanese students with varying English language proficiency levels and found that participants who had access to linguistically simplified or elaborated texts scored higher on the RC tasks than those who read the original text. Similarly, \citet{crossley2014s} showed a linear effect of text complexity on comprehension even when accounting for individual language and reading proficiency differences as well as their background knowledge.  \citet{rets2021simplify}'s study with 37 adult English L2 users showed better comprehension and faster recall for participants with low English proficiency levels with simplified texts. Finally, in a within-subject 
 study with four original and four simplified texts involving 103 participants with varying levels of English proficiency (beginner to native), \citet{temnikova-maneva-2013-c} show that utilizing Controlled Language \cite{temnikova-etal-2012-clcm} for TS improves reading comprehension. All these studies are conducted at the paragraph or the document level based on human-written simplifications which are implicitly assumed to be correct.

The use of reading comprehension for the evaluation of automatically simplified texts has been more limited. \citet{angrosh-etal-2014-lexico} first used reading comprehension to evaluate \textit{automatically} simplified texts from multiple \ts models, with non-native readers of English. They conduct a multiple-choice test using five news summaries chosen from the Breaking News English website, originally at reading level 6 (hard) and simplified manually or via automatic \ts systems. Their study found no significant differences between the comprehension accuracy of different user groups when reading automatically generated simplifications. However, they note that the drop in comprehension scores for some of these systems could be accounted to the content removal which can make some questions unanswerable. Hence, it is not clear whether the differences are non-significant due to user understanding, errors introduced by TS systems, or the effectiveness of simplifications. 
\citet{laban-etal-2021-keep} also conduct a reading comprehension study to evaluate the usefulness of automatic TS outputs with automatically generated questions that can be answered by original text and human-written references. They found that shorter passages generated by automatic TS systems lead to a speed-up in the RC task completion time regardless of simplicity. However, the automatic generation of questions mostly limited them to factoid, thus limiting the scope of understanding tests, and it is unclear whether the TS errors could render the RC questions unanswerable.

Evaluating text generation via automatic question answering (QA) has also received much attention, including for machine translation \cite{han-etal-2022-simqa}, and for text summarization, where it has been used to assess the factuality \cite{wang-etal-2020-asking} or faithfulness \cite{durmus-etal-2020-feqa} of model-generated summaries. For summarization evaluation, questions have been automatically created based on key information from the model-generated summary, such as important nouns or entities. An automatic QA system is then employed to generate answers to these questions using the original document as a reference. The quality of the generated summary is determined by comparing these answers using metrics that measure semantic similarity or exact matches. However, unlike summarization where the primary goal is to condense a text (either in an extractive or abstractive fashion), text simplification also involves making structural and linguistic changes so that the text is easier to comprehend which the existing automatic QA-based evaluations are not equipped to assess.

In this work, we design a reading comprehension task to assess the ability of \ts systems to preserve meaning via carefully constructed multiple-choice questions targeting language comprehension and use it to conduct a thorough controlled evaluation of a diverse set of state-of-the-art \ts systems. We conduct our human evaluation at the paragraph level as humans naturally tend to simplify complex text at this granularity, and utilizing complete texts for measuring RC would yield more accurate results compared to relying on individual sentences \cite{leroy2022evaluation}.

\section{A Reading Comprehension-based Human Evaluation Framework} \label{sec:method}

\paragraph{Overview} Our human evaluation is based on the following task: participants are presented with text and then are asked questions to test their understanding of some of the information conveyed in the text, as illustrated in Figure~\ref{fig:example_qa}. We seek to measure whether participants who read simplified versions of the original paragraph can answer questions as well as those who read the original. However, our goal is not to assess the participants but \ts systems: when working with participants who are proficient in the language tested, we assume that differences in reading comprehension accuracy indicate differences between the quality of \ts systems that produce the different simplifications. 

\paragraph{OneStopQA} Within this simple framework, the design of the RC questions and answers is critical to directly evaluate the correctness of automatic \ts systems. We build on the OneStopQA reading comprehension exercises created using the STARC (Structured Annotations for Reading Comprehension) annotation framework \cite{berzak-etal-2020-starc}, which is well suited to our task since it targets the real-world need of supporting readers with low English proficiency, and there is already evidence that it is a sound instrument to capture differences in reading comprehension from human-written text.

Specifically, OneStopQA is based on texts from the \url{onestopenglish.com} English language learning portal \cite{vajjala-lucic-2018-onestopenglish}, which are drawn from The Guardian newspaper. \sa{updated to add the nature of questions} Questions are designed to assess language comprehension rather than numerical reasoning or extensive external knowledge. More importantly, these questions can not be answered with simple string-matching and guessing strategies. Furthermore, the answer options under the STARC annotation framework follow a structured format that reflects four fundamental types of responses, ordered by miscomprehension severity: A indicates correct comprehension, B shows the ability to identify essential information but not fully comprehend it, C reflects some attention to the passage's content, and D shows no evidence of text comprehension \cite{berzak-etal-2020-starc}. 
Participants are presented with the answer options in a randomized order to minimize any potential bias or pattern recognition. The correct answer typically is not present verbatim in the critical span, a text span from the passage upon which the question is formulated.\footnote{We do not use the gold or distractor spans in the evaluation study or when generating the TS outputs. } We note that the questions only target a subset of the information conveyed in a passage, and hence, our evaluation framework does not provide a measure of completeness. In other words, correctly answering the RC questions does not require understanding every piece of salient information from the original. \footnote{70\% of the passages have critical spans (over the three questions) of at least 60\%, showing that the questions generally cover most information conveyed in the original text .} 

Further, prior work suggests that OneStopEnglish text and OneStopQA questions provide a sound basis for evaluating automatic \ts, as they can capture differences in reading comprehension from manually simplified text: \citet{gooding-etal-2021-predicting} found a statistically significant difference between users scrolling interactions and the text difficulty level in a 518-participant study and \citet{vajjala-lucic-2019-understanding} showed that the nature of the reading comprehension questions can impact text understanding. 

\paragraph{Targeting Answerability}  We augment the OneStopQA answer candidates with a fifth option motivated by the failure modes of automatic \ts. For each question, participants have the option to pick ``unanswerable'' (UA), which they are instructed to select when ``The questions or the answer options are not supported by the passage.''. This lets us directly measure how often readers judge that there is no support for answering the question based on the input text, which is a more salient problem when presenting participants with automatic than human-written simplifications. The resulting reading comprehension problems are illustrated in Figure~\ref{fig:example_qa}.

\paragraph{Text Granularity}  Participants are presented with a paragraph of text before answering each question, thus moving away from the prior focus on evaluating \ts at the sentence level. In real world settings, people are unlikely to use text simplification on isolated sentences and might be able to understand important information by making inferences from the context. Thus evaluating text simplification outputs at the paragraph level strikes a good balance between providing a realistic amount of context to readers without making the task too long.

\paragraph{Measures} 
Given $M$ paragraphs from one of the following: original text, human-written simplification, or the nine automatic TS systems, each paragraph $P \in M$, is accompanied by a set of $Q$ questions with the 5 multiple-choice answers $\{q, a_1^5\}_{1}^q$, as described above. We measure the adequacy of the simplified texts ($Acc$) using the number of questions answered correctly for that system by human participants. Formally, 

\begin{equation} \label{eq:rc_score}
   Acc = \frac{1}{M \times Q} \sum_{m=1}^M \sum_{q=1}^Q 1[ Selected == Correct]
\end{equation}

where Selected is the answer marked by human participants for a given passage $P$. We rank the automatic \ts systems based on the ranking induced by the above scores. Systems with higher scores produce simplifications that help people answer reading comprehension questions correctly.

\begin{table*}[t]
\centering
 \setlength\tabcolsep{2.5pt}
\scalebox{0.80}{
\begin{tabular}{lrrrccccrccr}
  \rowcolor{gray!30}
     \textbf{\textsc{MODEL}} & \multicolumn{3}{c}{\textbf{\textsc{TEXT}}} &&\multicolumn{2}{c}{\textbf{\textsc{CONFIG}}} &&\multicolumn{2}{c}{\textbf{\textsc{METRICS (P)}}} &&
     \\
      \rowcolor{gray!20}
      &
     \multicolumn{1}{c}{\textbf{\textsc{\# (Words}}} & \multicolumn{1}{c}{\textbf{\textsc{\# (Sents)}}} & \multicolumn{1}{c}{\textbf{\textsc{FKGL}}}   &
     &\multicolumn{1}{c}{\textbf{\textsc{Arch}}}  &\multicolumn{1}{c}{\textbf{\textsc{Data}}}  &&
     \multicolumn{1}{c}{\textbf{\textsc{SARI}}} & \multicolumn{1}{c}{\textbf{\textsc{BERTScore}}} && \\
    
\textsc{Original}  & 137.4 & 5.1 & 10.5 &&  - & - &&- & -&& P  \\
\textsc{Elementary} & 118.1 & 5.7 & 7.4 && - & -  &&- & - && P\\
\addlinespace[0.1cm]
\musssup & 126.8 & 7.3 & 7.0  && BART & \multirow{2}{*}{\textsc{WikiLarge}} && 45.07 & 0.940&& S  \\
\wiki & 135.0 & 7.7 & 6.6  && T5 && &  44.76 & 0.938 && S\\
\addlinespace[0.1cm]
\controlsup-Grade7 &  132.8 & 5.9 & 	9.0  && \multirow{4}{*}{\textsc{Transformer}}& \multirow{4}{*}{\textsc{Newsela}} && 29.27 & 0.946&& S\\
\controlsup-Grade5 &  124.1 &	7.3 &	6.8 && & && 38.35 & 0.939&& S \\
\editnar-Grade7 & 134.7 &	6.1 &	9.0  && & && 30.49 & 0.939&& S\\
\editnar-Grade5 & 131.0 &	8.3 &	6.1  &&& && 39.69 & 0.929&& S\\
\addlinespace[0.1cm]
\chatgpt &123.0 &	5.1 &	10.5  && GPT-3.5 & -  && 41.41 & 0.927 && P\\
\mussunsup & 124.7 &	5.7 &	9.1 &&{\textsc{BART}} & - && 40.67 & 0.937 && S\\
\kis & 73.6 &	3.3 &	9.1  && {\textsc{GPT-2}} & - && 33.06 & 0.893&& P\\
  \bottomrule
 \end{tabular}
}
\caption{Simplified texts are shorter and include more sentences than the Original. Automatic TS models use various architectures and datasets, generate simplified texts at either the sentence (S) or the paragraph (P) level, and show different tradeoffs in adequacy-simplicity (measured using BERTScore and SARI computed at the paragraph level (P)). A 0.005 difference in BERTScore is significant (p-value = 0.00).}  \label{tab:output-stats}
\end{table*}

We compute \textbf{answerability} ($Ans$) using questions that were marked ``UA'' by the participants:
\begin{equation} \label{eq:na_score}
   Ans = 1 - \frac{1}{M \times Q} \sum_{m=1}^M \sum_{q=1}^Q 1[ Selected == UA]
\end{equation}
Systems often produce outputs that do not support answering the question, perhaps due to over-deletion or other serious output pathology \cite{devaraj-etal-2022-evaluating}. Systems with high $Ans$ scores produce outputs that are not necessarily correct but still support answering the questions.

\section{Experimental Setup} \label{sec:setting}

First, we describe experiment details including data, participant selection, and study design. Then, we outline the selected \ts systems for evaluation.

\subsection{Study Design}
\label{subsec:data}

\paragraph{Data}  
The OneStopQA dataset includes 30 articles containing 162 paragraphs in total at three difficulty levels: Elementary, Intermediate, and Advanced. Each passage is accompanied by three multiple-choice questions that can be answered at all levels of difficulty.  The simplified versions include common text simplification operations such as text removal, sentence splitting, and text rewriting. We select the first two paragraphs from each of the 30 articles and associated questions resulting in 60 unique passages and 180 questions in total. Unlike prior studies that evaluate the impact of human-generated simplifications on various target audiences using only a limited number of articles (typically 1-5) and questions (around 3-5) \cite{long1993modifications,tweissi1998effects, oh2001two, crossley2014s, rets2021simplify}, our evaluation is on a larger scale (180 diverse passage-question pairs), which provides more statistical support to rank different systems. 

\paragraph{Participants} 
The participants are paid directly through the crowd-sourcing platform at an average rate of USD 15/hour. 
The task is conducted on the Prolific crowd-sourcing platform.\footnote{{\small \url{prolific.co}}}, we recruit 112 native speakers of English between ages 18-60 identified by their first language and with an approval rating of at least 80\% for evaluating the correctness of \ts systems. 

\paragraph{Task Design}
Each participant is provided with the following instruction: \textit{In this study, you will be presented with 6 short excerpts of English text, accompanied by three multiple-choice questions. You are asked to answer the questions based on the information presented in the text.}. A participant is presented with a random subset of 6 texts from one of the 11 conditions: original, simplified by humans, or simplified by one of the nine \ts systems.
Each passage-question pair is annotated by one native English speaker resulting in 1980 annotations. Annotations collected were manually spot-checked for straightlining (pattern where participants consistently select the same response option) and time differences to ensure that the participants were paying attention to the RC task.

\subsection{Models for Evaluation} We generate simplified outputs for the selected passages at ``Advanced'' difficulty, i.e., the Original text, using the systems described below as they are representative of the variety of architectures and learning algorithms (supervised, unsupervised, black-box) proposed in the \ts literature:

\begin{enumerate}
    \item Keep-it-simple (\kis) \cite{laban-etal-2021-keep} is an unsupervised \ts system trained using a reinforcement learning framework 
    to enforce the generation of simple, adequate, and fluent outputs at the paragraph level.\footnote{\url{github.com/tingofurro/keep_it_simple}}
    \item MUSS \cite{martin-etal-2022-muss} finetunes a BART-large \cite{lewis-etal-2020-bart} model with control tokens \cite{martin2020controllable} extracted on paired text simplification datasets and/or mined paraphrases to train both supervised and unsupervised \ts systems.\footnote{The control tokens are added to the beginning of the input acting as side constraints \cite{SennrichHaddowBirch2016c} and specify the text transformation, like the compression (via the length ratio between the source and the target) or degree of paraphrasing (via the character-level Levenshtein similarity).  Please refer to \citet{martin2020controllable} for more details.
    } 
    We use the suggested hyperparameters from the original paper to set the control tokens during simplification generation.\footnote{ {\small \url{github.com/facebookresearch/muss} }}
    \item ControlT5-Wiki \cite{sheang-saggion-2021-controllable} is a supervised controllable sentence simplification model that finetunes a T5-base model with control tokens. Again, we use the suggested hyperparameters from the original paper.\footnote{{\small \url{github.com/KimChengSHEANG/TS_T5}}}
    \item ControlSup \cite{scarton-specia-2018-learning} is a controllable supervised \ts model that trains a transformer-based sequence-to-sequence model with U.S. target grade as a side-constraint to generate audience-specific simplified outputs. We generate simplified outputs corresponding to Grades 7 and 5 to match the target complexity of the human-written Elementary simplified texts and to assess the impact of the degree of simplification on correctness.  
    \item EditingCurriculum (EditCL), proposed by \cite{agrawal-carpuat-2022-imitation} trains a supervised edit-based non-autoregressive model that generates a simplified output for a desired target U.S. grade level through a sequence of edit operations like deletions and insertions applied to the complex input text.  We generate simplified outputs corresponding to Grades 7 and 5.\footnote{{\small \url{github.com/sweta20/EditingCL}}}
    \item We generate paragraph-level simplified outputs using ChatGPT with the following prompt:\footnote{{\small \url{openai.com/blog/chatgpt}}}  \\
   {\small \texttt{\{Text\}}} \\
    {\small \texttt{Rewrite the above text so that it can be easily understood by a non-native speaker of English: }} 
\end{enumerate}

We also include the Elementary version of the text from the OneStopEnglish corpus to compare the reading comprehension of the original and model-generated simplified texts against a ground truth reference as a control condition.

Statistics for the human-written and automatically generated passages as well as model summary are presented in Table~\ref{tab:output-stats}. Automatically generated or manually written simplified texts are shorter and include more sentences (due to sentence splitting) than the original unmodified text. Systems that use pre-trained knowledge (\texttt{MUSS, T5, ChatGPT}) receive a higher simplicity (\sari) score than models trained from scratch (\texttt{ControlSup, EditCL}) except \texttt{KIS} that achieves low simplicity and adequacy scores according to automatic metrics.\footnote{\sari measures lexical simplification based on the words that are added, deleted, and kept by the systems by comparing system output against references and the input text.} Both \texttt{ControlSup} and \texttt{EditCL} models generate simplified outputs at a higher complexity level than intended (Average FKGL for Grades 7 and 5 are Grades 9 and 7 respectively). Furthermore, the outputs span a wide range of adequacy and simplicity scores where some systems trade-off adequacy for simplicity with low BERTScore and high SARI values (e.g. \texttt{ChatGPT, EditCL-Grade5}) and vice-versa (e.g. \texttt{ControlSup-Grade7}). While the range of BERTScore values appears small, differences of $>0.005$ are statistically significant suggesting that the 0.4+ wide range includes meaningful differences within this set of systems.

\section{Results} \label{sec:results}

We first analyze the results to show the validity of the evaluation set-up, before comparing \ts systems on the accuracy and answerability metrics.

\begin{table*}[ht]
   \centering
    \scalebox{0.85}{
    \begin{tabular}{llccccccrl}
    \toprule
 \textbf{\textsc{Type}}  & \textbf{\textsc{MODEL}}  & \multicolumn{1}{c}{\textbf{\textsc{Pre-trained}}}   & \textbf{\textsc{\% Correct}} & \textbf{B} & \textbf{C} & \textbf{D}   & \multicolumn{2}{c}{\textbf{\textsc{Rank}}}  \\
 \midrule
   \addlinespace[0.1cm]
\multirow{2}{*}{\textsc{Human}} & \textsc{Original} & - & 78.33  & 6.11 & 2.22 & 1.11 & \textbf{\texttt{1}} &  \\
& \textsc{Elementary}  & -  & 77.22  & 5.56 & 2.78 & 0.00 & \textbf{\texttt{2}} & \\
  \addlinespace[0.3cm]

\multirow{5}{*}{\textsc{Supervised}} & \musssup  & \cmark & 76.11 & 6.67 & 1.67 & 1.67 &\textbf{\texttt{3}}&  \\
& \wiki  & \cmark & 74.44   & 6.11 & 2.78 & 1.67 & \textbf{\texttt{4}}&  $*$\\
& \controlsup-Grade7  & \xmark   &  70.56  & 3.89 & 2.78 & 2.78 & \textbf{\texttt{7}}& \\
& \editnar-Grade7 & \xmark  & 69.44 & 10.56  & 2.22 & 0.56 & \textbf{\texttt{8}}  &$**$ \\
& \editnar-Grade5 & \xmark  & 69.44 & 10.00 & 2.22 & 0.00 & \textbf{\texttt{8}}&  $**$ \\
& \controlsup-Grade5 & \xmark  & 67.78 & 11.11 &  3.89 & 0.00 & \textbf{\texttt{10}} & \\

  \addlinespace[0.3cm]

\textsc{Black Box}  & \chatgpt  & \cmark & 74.44 & 9.44 & 1.11 & 0.00 & \textbf{\texttt{4}} & $*$ \\
  \addlinespace[0.3cm]
  
\multirow{2}{*}{\textsc{Unsupervised}} & \mussunsup  & \cmark &  73.33 &  6.67 & 2.78 & 1.11 & \textbf{\texttt{6}} &\\
& \kis  & \cmark & 20.50 & 7.22 & 3.89 & 3.89 & \textbf{\texttt{11}} &\\
  \bottomrule[1.5pt]
 \end{tabular}
}
\caption{Supervised systems that leverage pre-trained knowledge achieve the highest accuracy on the RC tasks. $*$ and $**$: Systems that attain the same rank due to the same overall accuracy scores.} \label{corectness-main}
\end{table*}


\subsection{Validity of the Human Evaluation} \label{subsec:validity}

\paragraph{Results on human-written texts align with the literature.} As can be seen in Table~\ref{corectness-main}, human-written texts (\texttt{Original, Elementary}) achieve the highest accuracy scores of approximately 78\%. This is consistent with a study by  \citet{berzak-etal-2020-starc} who report that Prolific crowd workers achieve a score of 80.7\% when tested on all 162 passages from OnestopQA. As expected, even with human written texts, participants do not answer all questions perfectly, reflecting individual differences in reading proficiency, background knowledge, and familiarity with the topic \cite{young1999linguistic, rets2021simplify} as well as the difficulty of the questions. These results thus provide an upper bound to contextualize the scores obtained by \ts systems. Furthermore, the scores obtained on the \texttt{Original}  and \texttt{Elementary} versions are very close, as expected when working with native speakers.

\begin{figure}[t]
\centering
\begin{subfigure}{\linewidth}
  \centering
\includegraphics[width=\textwidth]{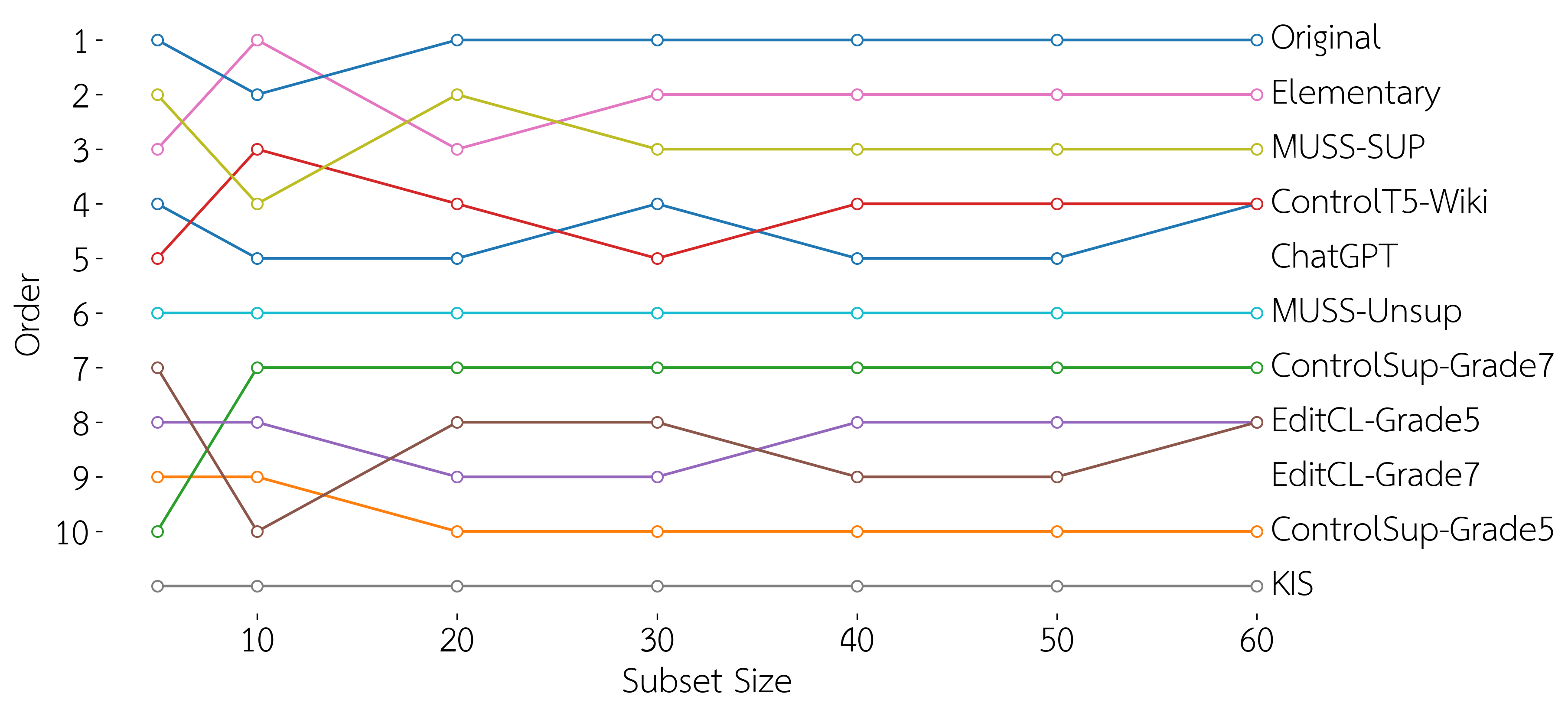}
\end{subfigure}%
 \caption{The accuracy scores per condition are averaged over $50$ runs for each subset-size $k$: Rankings stabilize with a sample size of 40 passages.}\label{fig:rc_score}
\end{figure}

\paragraph{Inter-annotator Agreement (IAA).} We collect a second set of annotations for a subset of 6 passages, covering all 11 conditions, and compute the IAA using Cohen’s kappa \cite{mchugh2012interrater}. The IAA score for selecting the correct answer indicates moderate agreement ($0.437$) despite the high subjectivity (individual comprehension differences) and complexity (5 answer options) of the task. 

\paragraph{System rankings are stable.} Sampling 50 random subsets of $k$ passages for $k \in \{5, 10, 20, 30, 40, 50, 60 \}$, we aggregate the mean accuracy score for each subset size and show the rankings for the systems in Figure~\ref{eq:rc_score}. Using $>40$ unique passages for each condition, i.e. approximately $120$ questions stabilizes the rankings amongst the 11 systems, with \texttt{ChatGPT, T5} and \texttt{\editnar-Grade7, Grade5} system pairs achieving the same rank. 

Taken together, these findings suggest that the evaluation framework is sound and provides a valid instrument to evaluate and compare systems.

\subsection{\ts Adequacy Findings} \label{sec:correctness}

\begin{figure*}[htb!]
\centering
\begin{subfigure}{0.95\linewidth}
  \centering
\includegraphics[width=\textwidth]{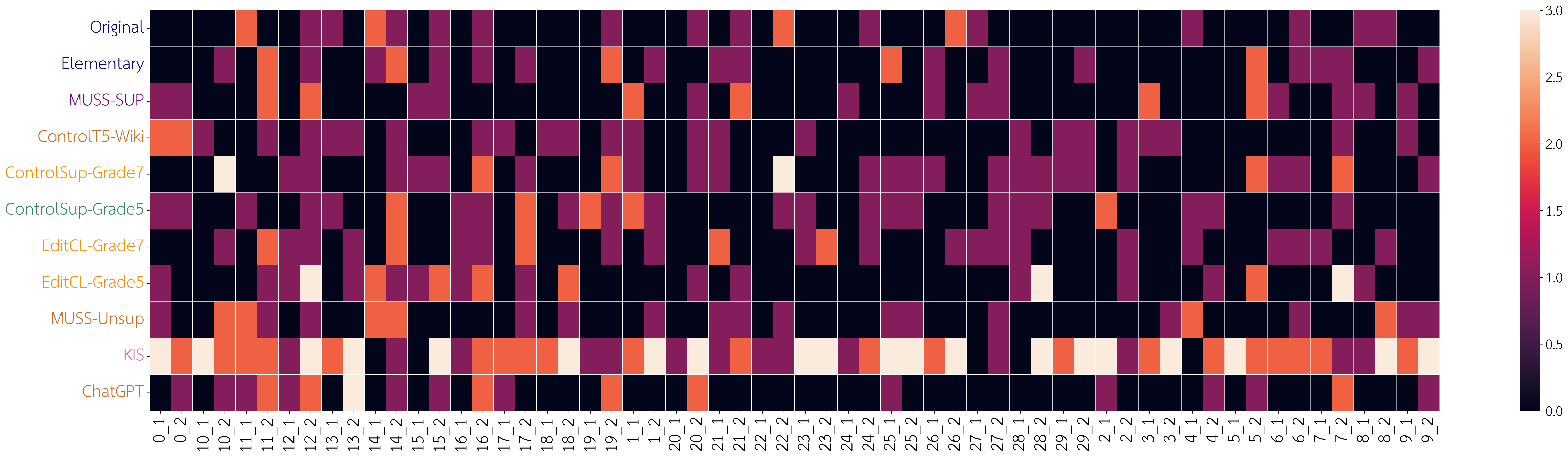}
\end{subfigure}%
 \caption{Number of questions marked with UA per paragraph: Different systems have different passage-questions pairs marked as unanswerable, suggesting different deletion errors.}\label{fig:heatmap_na}
\end{figure*}

Table~\ref{corectness-main} shows the $Acc$ scores for human-written texts (\texttt{Original, Elementary})
and automatic simplifications generated from supervised (edit and non-edit based), unsupervised, and black-box LLMs.
Systems achieve a wide range of scores, starting as low as 20\% to approaching within 1\% of the accuracy achieved on human-written text.  We discuss the main findings below.

Results first show that systems based on unsupervised pre-training yield more correct answers. This is the case for \texttt{\musssup} which achieves the highest accuracy among all systems.
\texttt{\textsc{ChatGPT}} attains a similar score to that of \texttt{\wiki}, a supervised sentence-level TS model, showing the benefits of large scale pre-training, and of reinforcement learning with human feedback \---\ even though it is unfortunately unknown whether \texttt{\textsc{ChatGPT}} was trained on TS or related tasks. Overall, the scores show that the best performing \ts systems rewrite content so that people understand the information tested as well as in human-written text. This suggests that those systems are worth including in usability testing in future work \---\ thus asking not only whether rewrites are \textit{adequate} as we do here, but also whether they are useful to readers that need simplified text.

At the other end of the spectrum, the texts simplified by \texttt{\kis} lead to answering only 20\% of questions correctly.  This is consistent with the low BERTScore for this system in Table~\ref{tab:output-stats}, and manual inspection which suggests that \kis is prone to deletions and hallucinations which do not preserve the meaning of the original. We will study the impact of deletions in more depth in the next section.

In the middle of the pack, among systems for grade-specific \ts, edit-based models outperform autoregressive models. The autoregressive model \texttt{ControlSup} exhibits a $3\%$ decrease in accuracy, due to a more aggressive deletion (Table~\ref{tab:output-stats}) when simplifying to Grade 5, whereas edit-based models like \texttt{\editnar} maintain their accuracy score even when generating simpler outputs at both grade levels 7 and 5. \rx{It would be good to analyze the B answers (or even C answers): instead of treating them as erroneous, is that a natural artifact of simplification?} \sa{Answered in response}However, these edit-based models also result in miscomprehension as suggested by the relatively high percentage of questions marked with option B by the human participants. Note that option B represents a plausible misunderstanding of the critical span upon which the question is based (Section~\ref{sec:method}). We hypothesize that this could be due to the reduced fluency of the model-generated simplifications via edit-based models. 

\begin{figure*}[htb!]
\centering
\begin{subfigure}{0.33\linewidth}
  \centering
\includegraphics[width=\textwidth]{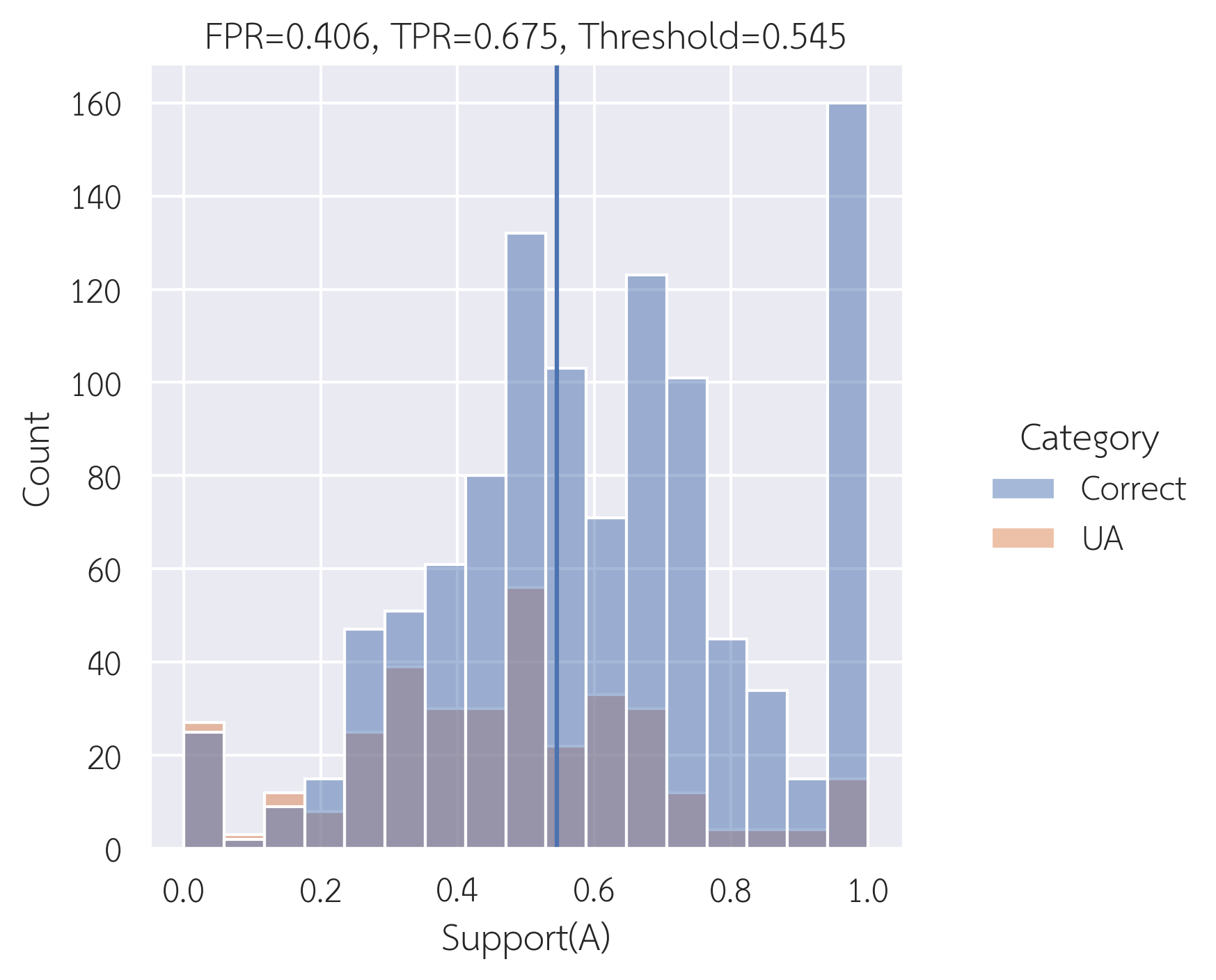}
\end{subfigure}%
\begin{subfigure}{0.33\linewidth}
  \centering
\includegraphics[width=\textwidth]{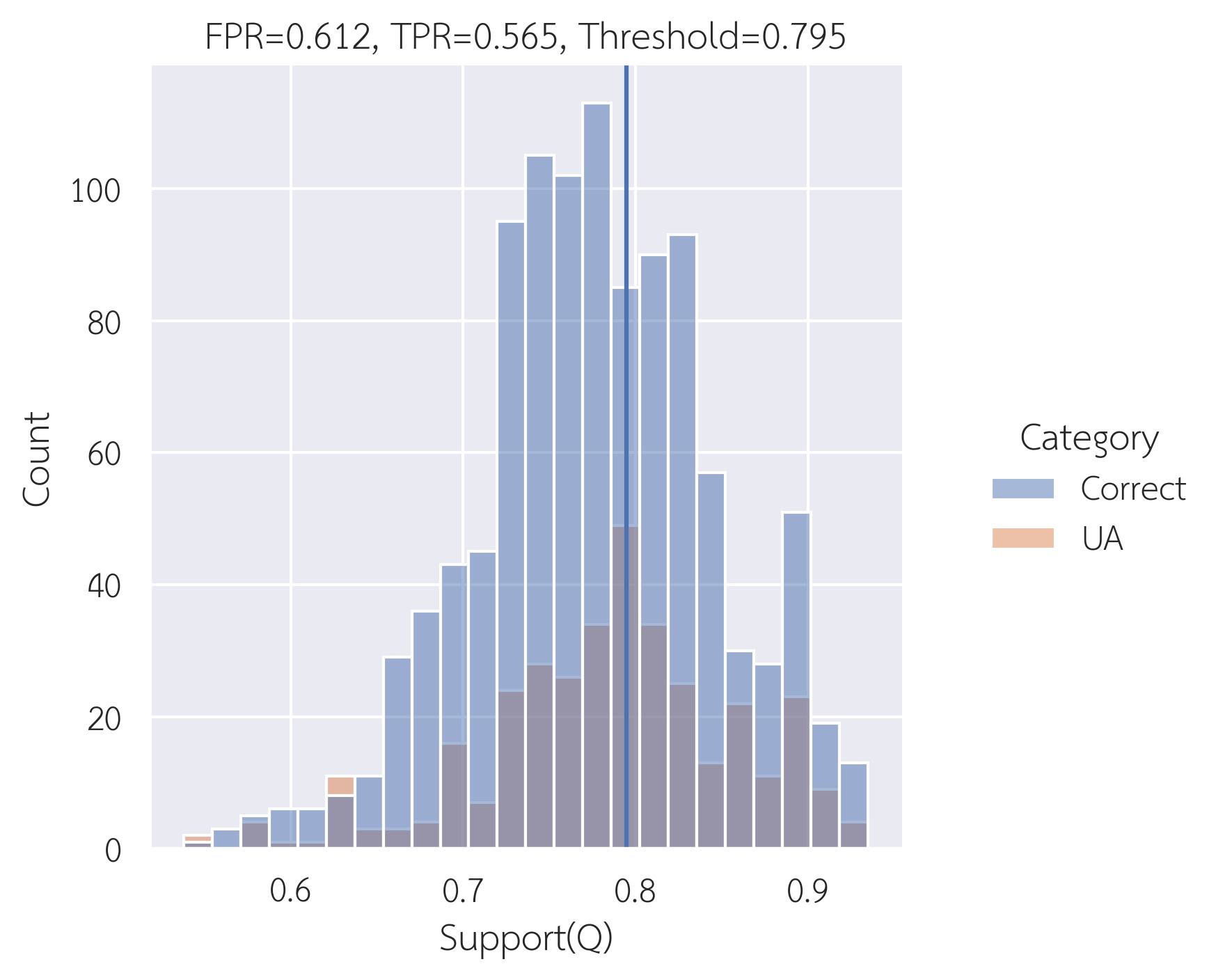}
\end{subfigure}%
\begin{subfigure}{0.33\linewidth}
  \centering
\includegraphics[width=\textwidth]{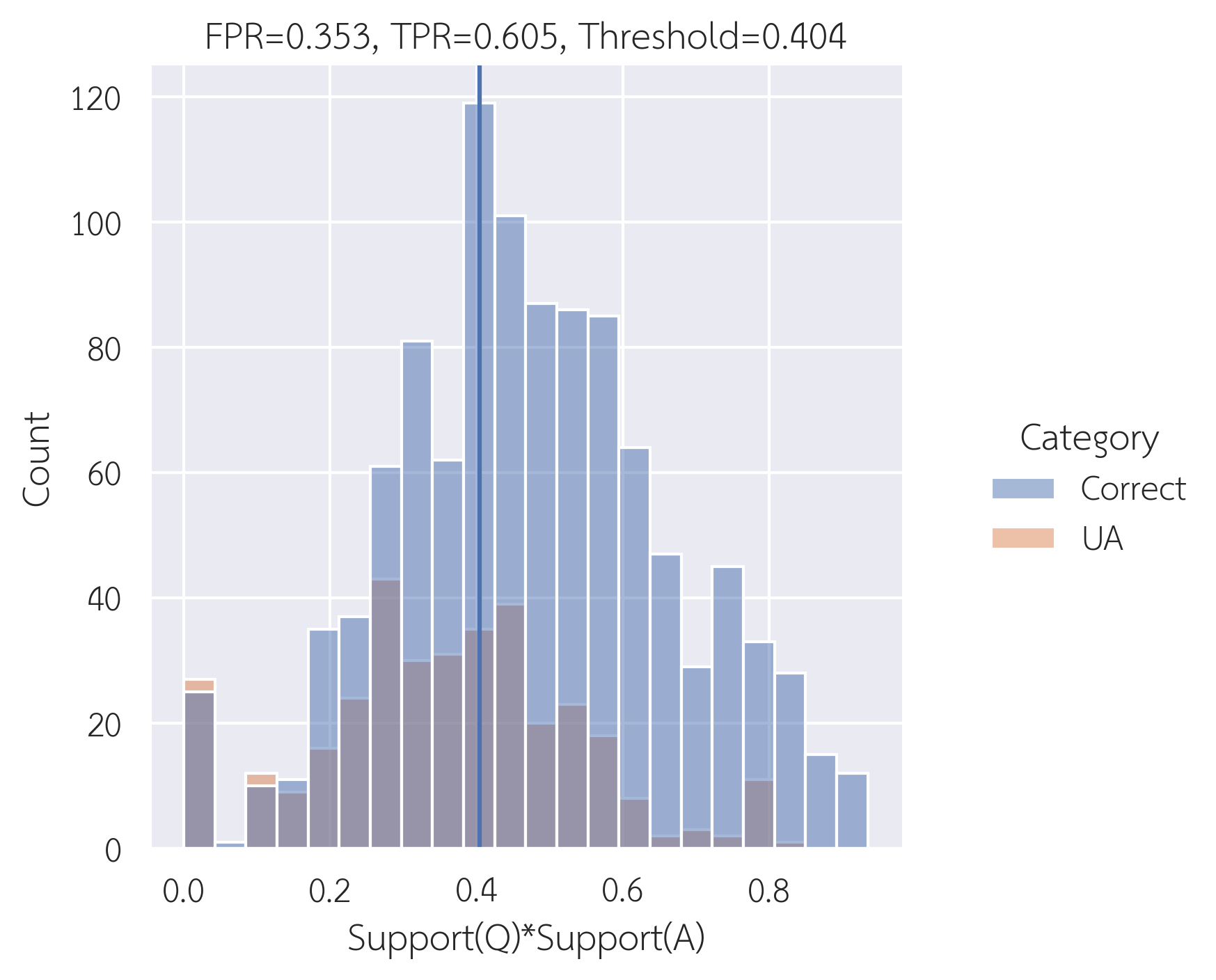}
\end{subfigure}%
 \caption{Unigram Overlap between the answer options and the passage (Support (A)), question and the passage (Support(Q)) and product of the two (Support (A)*Support (Q)) indicates content deletion as a major factor in making a question unanswerable. TPR: True Positive Rate; FPR: False Positive Rate.}\label{fig:support_ans_ques}
\end{figure*}

\subsection{TS Answerability Findings}
\label{sec:answerability}

\begin{figure}[ht]
\centering
\begin{subfigure}{\linewidth}
  \centering
\includegraphics[width=0.95\textwidth]{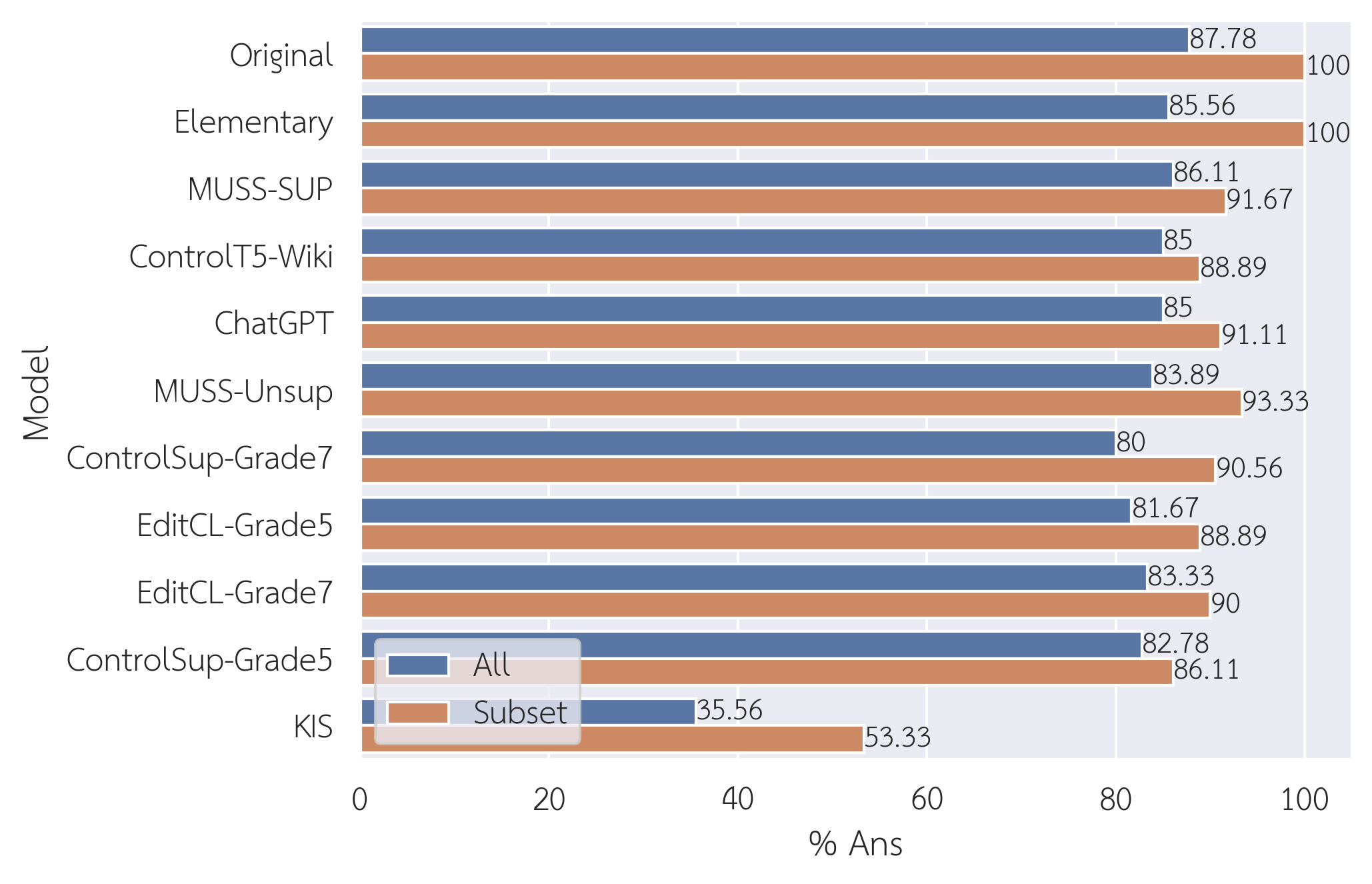}
\end{subfigure}%
  \caption{Participants mark 14-65\% questions with UA,  suggesting that the meaning of the original text is not entirely preserved in the simplified texts.}
  \label{fig:percent_na}
\end{figure}

 We show the answerability score, $Ans$, for all evaluation conditions in Figure~\ref{fig:percent_na}. 

Human-written text does not achieve perfect scores. Using the STARC annotation framework should ideally yield answerable questions, yet in practice, participants still mark 12-14\% of questions with UA. Manual inspection shows that these questions require making complex inferences and hypotheses about the plausibility of the various options. As a result, when given the UA option, participants are more conservative in selecting the four other alternatives.

Most systems achieve 83-86\% answerability for questions, except for \kis, which scores the lowest at 35.56\%. On the subset of questions answerable by both \texttt{Original} and \texttt{Elementary} texts, scores range from 53-92\%. This indicates that errors in model-generated texts hinder question answerability beyond individual comprehension differences. Models, except \kis, achieve similar scores but make different errors, as shown in Figure~\ref{fig:heatmap_na}, where no passage-question pairs are correctly answered by all models.

Building upon the finding of \citet{devaraj-etal-2022-evaluating} who show the prevalence of deletion errors in TS system outputs and our own manual inspection, we hypothesize that over-deletion is the key culprit that makes questions unanswerable. To test this hypothesis automatically, we examine how the unigram overlap (after stop word removal) between the question and the passage (\texttt{Support(Q)}) and the answer options and the passage (\texttt{Support(A)}) influence question answerability when model-generated outputs are used \cite{sugawara-etal-2018-makes}. While, in most cases, the correct answer does not appear as is in the critical span, we expect the unigram overlap to still provide a useful signal as the rephrased version often shares at least some unigrams with the critical span.

\begin{table*}[h]
\centering
\setlength\tabcolsep{2.5pt}
 \def\arraystretch{1.2}
\scalebox{0.68}{
\begin{tabular}{lrrrrrrrrrrrrrrrrr}
\toprule
    \multirow{3}{*}{\textbf{\textsc{metrics}}}  & \multicolumn{5}{c}{\textsc{\textbf{meaning (ref)}}} & \multicolumn{4}{c}{\textsc{\textbf{meaning (src)}}}  &   \multicolumn{4}{c}{\textsc{\textbf{simplicity}}} &  \multicolumn{1}{c}{\textsc{\textbf{readability}}}  & \multicolumn{2}{c}{\textsc{\textbf{QAFactEval}}} \\
     & \multirow{2}{*}{\bleu} & \multicolumn{3}{c}{\textsc{BERTScore}} & \multirow{2}{*}{\textsc{LevDist}} & \multicolumn{3}{c}{\textsc{BERTScore}} & \multirow{2}{*}{\textsc{LevDist}} & \multicolumn{4}{c}{\sari}  & \textsc{fkgl} \\ 
     & & (P) & (R) & (F1) & & (P) & (R) & (F1) & & (A) & (K) & (D) & (Avg.) &&  (F1) & (EM) \\
       \midrule
       \textbf{\textsc{All}}  & -0.193  & 0.418 & 0.292 &0.310 & -0.142 & 0.084 & 0.033 & 0.033 & -0.159 & 0.686 & -0.134 & 0.301 & \textbf{0.728}  & 0.126 & 0.126 & 0.025  \\
       \textbf{\textsc{All $-$ \{KIS\} }} & 0.157 & 0.167 & -0.012 & 0.012 & -0.634 & -0.311 & -0.383 & -0.383 &-0.659 & 0.719 & -0.622 & 0.707 & \textbf{0.778}  & 0.335 & -0.252 & 0.395 \\ 
  \bottomrule
 \end{tabular}
}
\caption{Evaluation of Automatic Metrics computed at the Paragraph-level using All (9) and All but KIS (8) automatic \ts systems: SARI achieves the highest correlation with human judgments, followed by LevDist, surpassing meaning preservation and readability metrics: BLEU, BERTScore, and FKGL. } \label{tab:automatic-main}
\end{table*}

Figure~\ref{fig:support_ans_ques} shows that \texttt{Support(A)}, is a more reliable predictor of UA with a true positive rate (TPR) of 0.675 at a false positive rate (FPR) of 0.406 than \texttt{Support(Q)} (TPR: 0.565, FPR: 0.612). Answers that appear verbatim in the passage (Support(A)=1.0) are correctly answered 93\% of the time. However, when the question lacks support in the passage, the unigram overlap with just the answer becomes an insufficient signal. Therefore, we also report the distribution of the product of \texttt{Support(A)} and \texttt{Support(Q)} in the same figure to directly capture the support for both the question and the answer options in the passage, i.e., the UA option. The resulting TPR rate for predicting UA is 0.605 at an FPR of 0.353, indicating that the incorrect deletion of partial or complete phrases by the systems affects the support for both the question and the answer options making RC question unanswerable.

These results temper the \correctness results, suggesting that even the best-performing systems delete content. Taken together these findings call for more research on calibrating the deletion tendencies of \ts systems, and for human subject studies to develop machine-in-the-loop workflows to validate automatically simplified content before it is presented to readers.

\section{Evaluating Automatic TS Evaluation Metrics} \label{sec:automaticmetrics}
We now turn towards investigating to what extent automatic \ts evaluation metrics frequently used in the literature capture the system rankings obtained via the RC task \cite{Thanyyan2021Data, maddela2022lens, devaraj-etal-2022-evaluating}.
 We compute the Spearman-Rank correlation of the system-level scores using selected automatic metrics and the RC accuracy scores in Table~\ref{tab:automatic-main}.  For meaning preservation, we evaluate \textsc{BLEU}, BERTScore \cite{BERTScoreZhang2020} and the Levenshtein distance computed between the system output and the Elementary text (\textsc{Ref}) or the system output and the Original text (\textsc{Src}). For simplicity and readability dimensions, we report correlation scores with SARI, and FKGL respectively. SARI measures lexical simplicity based on the n-grams that are kept (K), added (A), and deleted (D) by the system relative to the original text and to the reference simplified (elementary) texts. Note that all metrics are computed at the paragraph level, just like in the RC task, unlike prior evaluation which uses and evaluates these metrics for sentence-level simplification. We also report the correlation scores with QAFactEval, a QA-based metric designed to evaluate factual consistency in summaries \cite{fabbri-etal-2022-qafacteval}. \footnote{\url{https://github.com/salesforce/QAFactEval}}

Overall, \textsc{SARI} achieves the best correlation across the board with or without including the outlier system, i.e. \texttt{KIS}. The addition component (A) of SARI that rewards the insertion of n-grams present in the simplified reference but absent from the original text achieves a moderate-high correlation score (0.686-0.719) in both settings.
The Levenshtein edit distance of the system output with the \texttt{Original} (-0.659) and the \texttt{Elementary} (-0.634) text receives a negative moderate-high correlation with human judgments, outperforming both surface-form (BLEU) and embedding-based metric (BERTScore) after removing the outlier system (\kis). We hypothesize that metrics that focus on similarity to only the original or the simplified text do not fully capture the balance between simplicity and adequacy. SARI's 3-way comparison between the input, the output, and the reference is key in yielding system rankings that are consistent with those based on our accuracy results, which could be further repurposed to more directly align evaluation metrics with the accuracy scores.  

Furthermore, QAFactEval exhibits only a weak correlation (0.395) at best with human judgments. This is consistent with the current findings by \citet{kamoi-etal-2023-shortcomings} who discuss and show how automatically extracting facts from summaries could lead to a fundamental problem in the evaluation where current QA-based frameworks not only struggle to accurately identify errors in the generated summaries but also perform worse than straightforward exact match comparisons.

\section{Model-based Question Answering} \label{sec:modelqa}

Our evaluation so far has relied on human-written questions answered by crowd workers, using either model-generated or human-written texts. Automating one or both components would help scale the evaluation and port it to new settings more flexibly.  Recent work suggests that this might be plausible: \newcite{krubinski-etal-2021-just} show that automatically generated questions and answers can be used to evaluate Machine Translation systems at the sentence level, and automatic QA techniques \cite{fabbri-etal-2022-qafacteval, wang-etal-2020-asking} have been used to assess the factuality and faithfulness of summarization systems. 

\begin{figure}[h]
\centering
\begin{subfigure}{0.92\linewidth}
  \centering
\includegraphics[width=0.90\textwidth]{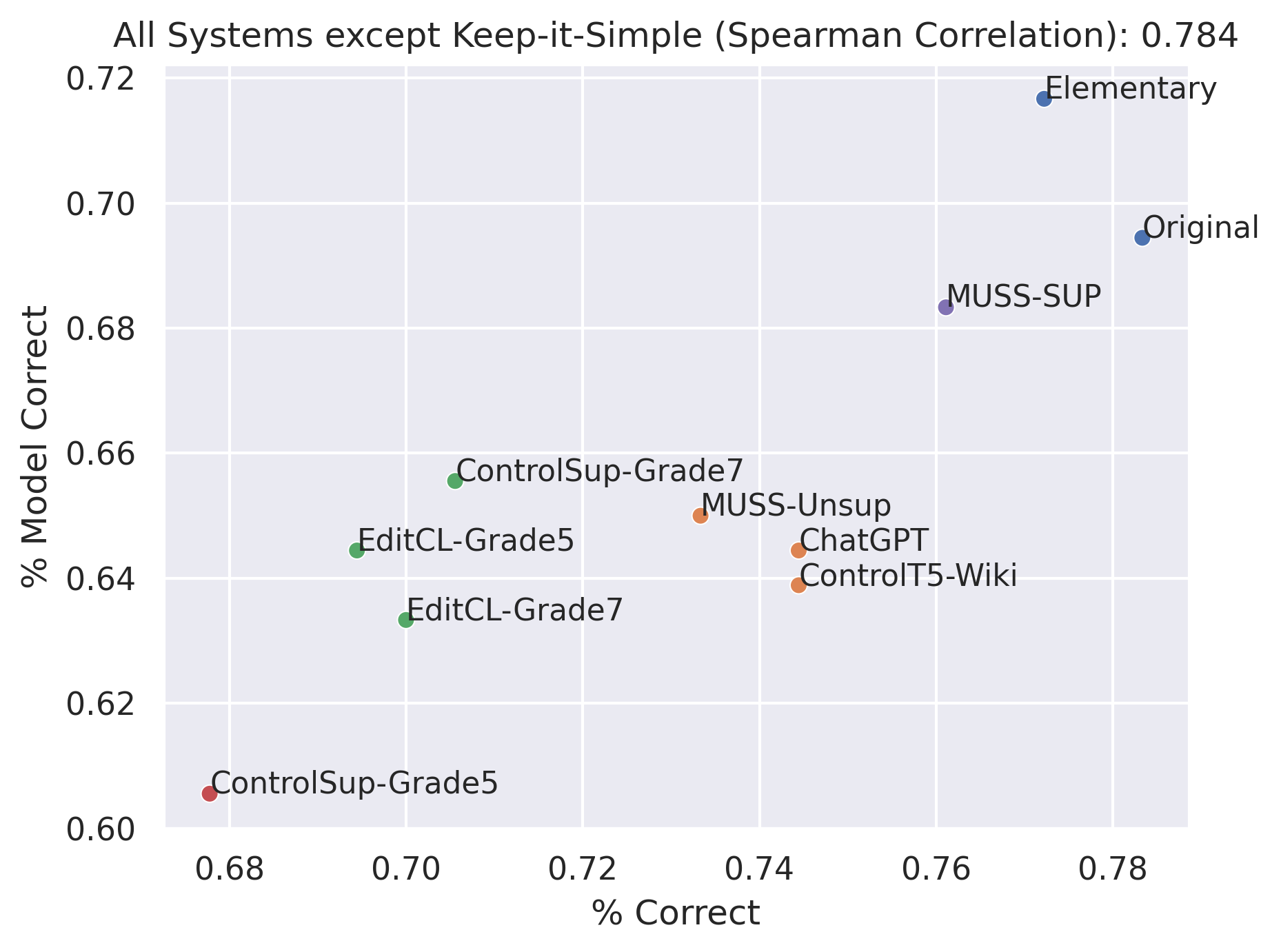}
\end{subfigure}%
\caption{Model-based QA achieves moderate-high correlation with human judgments but fails to distinguish closely competing systems. } \label{fig:system_model_correlation}
\end{figure}

Here, we assess the performance of a state-of-the-art QA system in recovering the gold-standard ranking induced by human judgments, leaving the more complex study of multiple-choice RC question generation to future work. We use \texttt{UnifiedQA v2} \footnote{\texttt{allenai/unifiedqa-v2-t5-3b-1363200}} a QA model, trained to answer questions in 4 different formats using 20 different datasets. This model has been shown to support better generalization to unseen datasets compared to models specialized for individual datasets. We use the format recommended in the original paper: \texttt{ \{question\} \textbackslash  n (A) \{choice 1\} (B)  \{choice 2\} ... \textbackslash n  \{paragraph\} } to generate answers for all the conditions. The Spearman-rank correlation between Exact Match (EM) and ground truth accuracies (C) for all systems is 0.838 and 0.744, excluding \kis. However, we note that the system's ability to distinguish closely competing systems (highlighted by the same color) is limited, as shown in Figure~\ref{fig:system_model_correlation}.

Interestingly, QA using human-simplified text achieves higher accuracy than using original unmodified text.  This finding is in line with prior work where \ts has been shown to improve the performance of multiple downstream \nlp tasks such as information extraction \cite{miwa-etal-2010-entity, schmidek-barbosa-2014-improving}, parsing \cite{chandrasekar-etal-1996-motivations}, semantic role labeling \cite{vickrey-koller-2008-sentence}, machine translation \cite{GerberHovy1998,SanjaStajnerMajaPopovic2016, HaslerdeGispertStahlbergWaiteByrne2017, stajner-popovic-2018-improving, miyata2019evaluating, mehta2020simplify}, among others \cite{van-etal-2021-may-help}. This suggests that automating part of the evaluation framework is a direction worth investigating in more depth in future work.

\section{Conclusion}

We introduced an evaluation framework based on reading comprehension to directly assess whether \ts systems correctly convey salient information from the original texts to readers. This framework lets us conduct a thorough human evaluation of the adequacy of 10 simplified texts: a human-written version and outputs from nine \ts systems.

Supervised systems that leverage pre-trained knowledge (MUSS, T5) produce texts that lead to the highest reading comprehension accuracy, approaching the scores obtained on human-written texts. Prompted LLMs (ChatGPT) perform well but are not as accurate as supervised systems. However, we find that even those systems do not preserve the meaning of the original text, with at least 14\% of questions marked as ``unanswerable'' on the basis of the text they generate. 

When human evaluation is not practical, our analysis suggests that SARI is a better metric than meaning-preservation metrics such as BERTScore and BLEU to rank systems by \correctness, and that model-based QA can approximate system rankings but at the cost of reduced discriminative power across systems and can introduce other confounding factors.

Overall, these results confirm the importance of directly evaluating the accuracy of the information conveyed by TS systems, and suggest that while some systems are overall correct enough to warrant usability studies, all systems still make critical errors. This motivates future work on machine-in-the-loop workflows to let editors and readers rely on \ts appropriately \cite{leroy2022evaluation}, and on improving the over-deletion of content by current \ts systems.  Our human evaluation framework provides a blueprint for evaluating whether correct \ts outputs improve reading comprehension for people who have difficulty understanding complex texts, which we intend to investigate in future work.

\section*{Acknowledgments}

We thank our TACL action editor, the anonymous reviewers, and the members of the UMD CLIP lab for their helpful and constructive comments on the paper. We also want to thank J. Jessy Li, Ani Nenkova, Philip Resnik, Jordan Boyd-Graber and Abhinav Shrivastava for their feedback on the earlier versions of the work. This research is supported in part by the Office of the Director of National Intelligence (ODNI), Intelligence Advanced Research Projects Activity (IARPA), via the HIATUS Program contract \#2022-22072200006, the NSF grant 2147292, funding from Adobe Research, the Portuguese Recovery and Resilience Plan through project C645008882-00000055 (Center for Responsible AI) and by Fundação para a Ciência e Tecnologia through contract UIDB/50008/2020. The views and conclusions contained herein are those of the authors and should not be interpreted as necessarily representing the official policies, either expressed or implied, of ODNI, IARPA, or the U.S. Government. The U.S. Government is authorized to reproduce and distribute reprints for governmental purposes notwithstanding any copyright annotation therein.

\bibliography{anthology,custom}
\bibliographystyle{acl_natbib}

\end{document}